\documentclass{elsart}
\usepackage{epsf}
\usepackage{graphics}
\usepackage{graphicx}
\usepackage{amssymb}
\usepackage{latexsym}
\usepackage{mathrsfs}
\usepackage{url}

\begin{document}
\begin{frontmatter}

\title{An Exact Algorithm for the Stratification Problem with Proportional Allocation}
\author{Jos\'e Brito}
\address{IBGE, Instituto Brasileiro de Geografia e Estat\'istica.Comeq, 20031-170, Rio de Janeiro, Brazil  
\\ {\tt jose.m.brito@ibge.gov.br}}
\author{Maur\'icio Lila}
\address{IBGE, Instituto Brasileiro de Geografia e Estat\'istica.Comeq, 20031-170, Rio de Janeiro, Brazil 
\\{\tt mauricio.lila@ibge.gov.br}}
\author{Fl\'avio Montenegro}
\address{IBGE, Instituto Brasileiro de Geografia e Estat\'istica.Comeq, 20031-170, Rio de Janeiro, Brazil 
\\{\tt flavio.montenegro@ibge.gov.br}}
\author{Nelson Maculan}
\address{COPPE, Universidade Federal do Rio de Janeiro, P.O. Box
  68511, 21941-972 Rio de Janeiro, Brazil \\ 
  {\tt maculan@cos.ufrj.br}} 

\begin{abstract}
We report a new optimal resolution for the statistical stratification problem under proportional sampling allocation among strata. Consider a finite population of \textit{N} units, a random sample of \textit{n} units selected from this population and a number \textit{L} of strata. Thus, we have to define which units belong to each stratum so as to minimize the variance of a total estimator for one desired variable of interest in each stratum, and consequently reduce the overall variance for such quantity. In order to solve this problem, an exact algorithm based on the concept of minimal path in a graph is proposed and assessed. Computational results using real data from IBGE (Brazilian Central Statistical Office) are provided. \\ [0.7em]
\end{abstract}
\end{frontmatter}


\section{Introduction}

A common procedure in sampling surveys is partitioning the elements of a population, before distributing the sample on it, in such a way to obtain most useful information from the data to be collected. This procedure is called stratification. It may have different aims, such as to guarantee obtaining information for some or all the geopolitical regions of a country, or to provide more precision in estimating population quantities by identifying strata with more homogeneous elements into them, according to one or more variables. In this latter case, the stratification is also called statistical stratification.

A principal use of statistical stratification, in order to obtain a better precision, is in defining what percentage of the sample must be taken from each stratum once we have chosen a non-uniform allocation scheme, that is, a non-trivial functional relation between the size of each stratum and the number of sample units to be collected in it. Thus, it is important to consider the allocation scheme itself in order to do a suitable statistical stratification.

In this paper, we propose an exact algorithm to solve the statistical stratification problem, that we call simply stratification problem, considering a simple non-uniform allocation scheme. Specifically, this method intends to solve the problem of optimal stratification with stratified simple random sampling without replacement \cite{Cochran} using proportional allocation. In this problem, we must divide a population of size $N$ into $L$ strata considering an auxiliary variable $X$, also called the size variable, whose values are known for all units in the population. The first stratum is defined as the set of units in the population whose $X$  values are lower than or equal to a constant value $b_1$ , the second one as the set of units whose $X$ values are greater than $b_1$  and lower than or equal  to $b_2$  and so on. Based on this definition the stratum $h$ $(h=2,...,L-1)$ is defined as the set of units in population with values of $X$ belonging to the interval $(b_{h-1} ,b_{h} ]$, where  $b_{1} < b_{2} <...<b_{L-1} $ are the boundaries of each stratum, and the stratum $L$ corresponds to the set of observations which values are greater than $b_{L-1} $. The problem of optimal stratification consists in to find boundaries $b_{1} <b_{2} <...<b_{h} <...<b_{L-1} $ which minimize the variance of the estimator of total for one or more variables $Y$ of study that are supposed to have some correlation with the $X$ variable, or even the $X$ variable properly. Aiming to solve this problem, a new algorithm, when proportional allocation is used, is proposed using the idea of minimal path in graphs \cite{Ahuja}.

This paper is organized in five sections. In section 2, we present some basic concepts about stratified simple random sampling. In section 3, we define the problem of stratification to be tackled in this work and offer a brief discussion about different approaches to this topic. We propose, in section 4, an algorithm based on Graph Theory in order to provide exact solutions to the stratification problem defined in section 3. Finally, we present some computational results and considerations about the new algorithm.

\section{A Review of Stratified Simple Random Sampling}

In stratified sampling  \cite{Cochran}, a population with $N$ units is divided into $L$ groups with $N_{1} ,N_{2} ,...,N_{h} ,...,N_{L} $ units respectively. These groups are called strata. There is no overlap among them and together they exhaust the population. Thus, we have

\begin{equation}
N_{1} +N_{2} +...+N_{h} +...+N_{L} =N
\end{equation}

After the strata definition, which is based on characteristics of the population, sampling units are selected in each sstratum, independently, according to a specific criteria of selection. The sample sizes of the strata are denoted by  $n_{1} ,n_{2} ,...,n_{h} ,...,n_{L} $, respectively.

Some basic notation about stratified sampling is presented as follows.

   $N$ -   Number of  units in the population ;

  $N_{h} $ -   Number of units in the stratum $h$ ;

    $n$ -    Total sample size;

   $n_{h} $ -   Sample size in the stratum $h$;

  $Y_{hi} $  -   Value of  a variable of interest $Y$ for the unit   $i$  in the stratum $h$;

 $\overline{Y}_{h} =\frac{\sum_{i=1}^{N_{h}} y_{hi}}{N_h} $    - Population mean of the variable  \textbf{\textit{Y}}  in the stratum \textit{h};

$Y=\sum _{h=1}^{L}N_{h} .\overline{Y}_{h}  $ - Population total of the variable \textbf{\textit{Y}};

$S_{hy}^{2} =\frac{\sum_{i=1}^{N_{h} }(y_{hi} -\overline{Y}_{h}  )^{2} }{N_{h} -1} $   - Population variance of  the\textit{ }variable \textbf{\textit{Y}}  in the stratum \textit{h};

\section{Problem Definition}

Consider a population $U=\{ 1,2,...,N\} $  with a study variable $Y_{U} =\{ y_{1} ,y_{2} ,...,y_{N} \} $ and an auxiliary size variable $X_{U} =\{ x_{1} ,x_{2} ,...,x_{N} \} $, $x_{1} \le x_{2} \le ...\le x_{N} $ , to be used in the stratification process \cite{Cochran}. A sample is taken from $U$ in order to estimate the population total $\widehat{Y}$ (where the symbol  ``\^{} `` stands for the estimated value of a population variable).

Suppose that the population must be divided, taking into account the values $X_{U} $, into $L$ disjoint groups, denoted by  $U_{1} ,U_{2} ,...,U_{h} ,...,U_{L} $.  The union of these $L$  groups  correspond to the complete indexes of the population $U_{1} \bigcup  ...\bigcup U_{L} =U$.       

According to  definitions in section 2, $\;N_{h}\;$   is the total of units in the population in each stratum $h$, and $n_{h} $ is the number of sampling units to be selected in the population in each stratum $h$, $h=1,2,...,L$. Thus, the set of indexes $U_{h} $, where $h=1,2,...,L$, contains the population indexes in each strata, such as:  $U_{1} =\{ i:x_{i} \le b_{1} \} $;   $U_{h} =\{ i:b_{h-1} <x_{i} \le b_{h} \} $, where $h=2,3,...,L-1$ ; and  $U_{L} =\{ i:b_{L-1} <x_{i} \} $, where $b_{1} <b_{2} <...<b_{h} <...<b_{L-1} $ denote the boundaries of each stratum in the population. 

We must emphasize that repeated observations associated to the population vector $X_{U} $ have to belong to the same stratum.

Let the sample size  $n_{h} $  and the population size $N_{h} $ , defined in each stratum, considering certain variable of interest \textit{Y}. The estimator \cite{Cochran} of the population total $\widehat{Y}$ is defined as follows: 

\begin{equation}
 \widehat {Y} =\sum_{h=1}^{L} \frac{N_h}{n_h} \sum_{k=1}^{n_h} y_{hk}
\end{equation}

  Then, the problem is to find strata boundaries $b_{1} <b_{2} <...<b_{h} <...<b_{L-1} $ that minimize the variance associated to the estimator of total $\widehat{Y}$, when simple random sampling without replacement in each stratum is adopted. The variance of $\widehat {Y}$ is defined as follows:
  
\begin{equation}
\label{eq:Variance}
  V(\widehat {Y}) =\sum_{h=1}^{L} {N_h}^2. \frac{S^2_{hy}}{n_h} . (1-\frac{n_h}{N_h})
\end{equation}

Notice that both  $N_{h} $  and $S_{hy}^{2} $ are defined according the strata's boundaries while $N$ is fixed previously. The sample sizes $n_{h} $ can be established using the following equations:

\begin{equation}
\label{eq:Proportional}
n_h = \frac{n.N_h}{N}
\end{equation}

\begin{equation}
\label{eq:Neyman}
n_h = \frac{n.N_h.S_{hy}}{\sum_{l=1}^{L}N_l.S_{ly}}
\end{equation}

 The equation \ref{eq:Proportional}  is associated to proportional allocation and (\ref{eq:Neyman}) to Neyman's allocation  \cite{Cochran}. Replacing $n_{h} $ in  (\ref{eq:Variance})  by (\ref{eq:Proportional}),  we have: 
 
 \begin{equation}
 \label{eq:Varianc2}
  V(\widehat {Y}) = \frac{N}{n} . (1-\frac{n}{N}) \sum_{h=1}^{L} {N_h}. S^2_{hy}
 \end{equation}

We have to emphasize that to find a global minimum for the variance described in  (\ref{eq:Varianc2}) considering proportional allocation or Neyman's allocation is a hard task to be done, either analytically or by intensive computing methods, because $S_{hy}^{2} $ is a nonlinear function of  $b_{1} <b_{2} <...<b_{L-1} $  and the number of possibilities of different choices for these values may be very high, as we shall see in section 4.

Because of that, several methods which yield a local minimum have been suggested. A well-known method of strata definition was proposed in \cite{Dalenius}. This method consists in approximating the distribution of the variable of stratification $X$  in the population using an histogram with various classes, which implies in adopting the hypothesis that the variable of stratification has an uniform distribution \cite{Cochran} in each class. In this case, the problem of stratification has an ordinary solution when the Cumulative Root Frequency Algorithm, or Dalenius-Hodges rule,  whose description can be find in \cite{Cochran}  (chapter 5), is applied.

In recent studies, Lavallée and Hidiroglou (\cite{Lavallee}) adopted a sample design using power allocation. The method aims to find stratification boundaries establishing the number of strata and the desired level of precision to minimize the overall sample size $n$.

According to Hedlin \cite{Hedlin}, the strata delimitation is considered such that the variance of the total estimator of a variable of interest, given by (\ref{eq:Variance}), has to be a minimum, considering  $n$ and  $L$  fixed previously and applying Neyman's allocation in each strata.

Gunning and Horgan \cite{Gunning} developed an algorithm that is easier to be implemented which applies the general term of a geometric progression to establish the boundaries of the strata when skewed populations are considered. This algorithm considers the hypothesis that the coefficients of variation are equal for all strata.

As we can observe, there are a number of methods applicable to the optimal stratification problem using Neyman or power allocation. However, at this moment no previous algorithm was proposed considering proportional allocation.

\section{Methodology}

In order to implement the new methodology, we had to modify the input data structure. Since the $N$ observations $X_{U} $ are ordered ascendingly, it is possible to gather them, taking into account only their distinct values. We have $K$ distinct values of $X_{U} $, gathered in a set $Q=\{ q_{1} ,q_{2} ,...,q_{K} \} $, which are the eligible boundaries to stratify the whole population.

Consider, for example, $N=9$,  $L=2$  and $X_{U} =(2,4,4,8,10,10,10,15,15)$. Then we have $Q=(2,4,8,10,15)=(q_{1} ,q_{2} ,q_{3} ,q_{4} ,q_{5} )\Rightarrow $ $|Q|=5=K$ and $I=\{ 1,2,3,4,5\} $, where  \textit{I  }is the set of indexes associated to the elements of  $Q$.

Defining $U_{1} =\{ i|x_{i} \le q_{3} ,x_{i} \in X_{U} \} $  and  $U_{2} =\{ i|q_{3} <x_{i} \le q_{5} ,x_{i} \in X_{U} \} $,  we have the following results:

\begin{table}[!ht]
\begin{center}
{\scriptsize
\begin{tabular}{p{0.6in} p{1.0in} p{1.2in} p{1.0in} p{1.1in}} \hline 
\textit{Stratum} & \textit{Population Size} & \textit{ Labels for each \newline       Stratum } & \textit{Observations \newline of $X_{U} $} & \textit{Indexes of Set  I} \\ \hline 
1 & $N_{1} =4$ & $U_{1} =\{ 1,2,...,4\} $ &      $\{ x_{1} ,...,x_{4} \} $ & $\{ 1,2,3\} $ \\ 
2 & $N_{2} =5$ & $U_{2} =\{ 5,...,9\} $ &      $\{ x_{5} ,...,x_{9} \} $ & $\{ 4,5\} $ \\ \hline 
\end{tabular}
}
\end{center}
\caption{Example of stratification}
\end{table}

Analogously, for $L$ strata  and $K$ boundaries , we have to find $L-1$  boundaries $q_{k} $ from $Q$ . Considering a finite population of size $N$, that will be divided in $L$ strata, and the ordered values $X_{U} $, we establish the sets $Q$ and  $I$. The solution for this problem will then consist of boundaries $q_{k} $ , selected from $Q$, or indexes \textit{i} , selected from $I$, that give the minimum variance according to (\ref{eq:Varianc2}).

The resolution of the problem above can be obtained by considering the enumeration of all the possible divisions of the observations associated to the set $I$, that is, by evaluating the variances of all the solutions and taking out the solution with the lower variance. However, this procedure may take an excessively high computational time even for moderately high values of the number of observations of $I$ and/or the number $L$ of strata. In fact, determining the number of solutions to be considered corresponds to the following combinatorial problem:

Determine the number $m$ of non-negative integer solutions of the equation  

\begin{equation} 
\label{eq:equation}
z_{1} +z_{2} +...+ z_{h} + ...+ z_{q} =r 
\end{equation} 

In this equation, each  $z_{h} $ corresponds to the number of elements of  $I$ allocated in each stratum \textit{h}, $q$ corresponds to the number $L$ of strata and $r$corresponds to the total of elements in $I$. In accordance with the stratification problem, it is necessary to make a little modification in equation (\ref{eq:equation}) in order to guarantee that the number of population observations is greater than or equal to 2 in each stratum ($z_{h} \ge 2$).

Performing a change of variable $z_{h} =t_{h} +2$, $t_{h} \ge 0$, and considering $q=L$ and $r=|I|$,  the equation (\ref{eq:equation}) can be rewritten as

\begin{equation} 
\label{eq:Sumation} 
t_{1} +t_{2} +...+ t_{h} + ...+ t_{L} =|I|-2L 
\end{equation} 

It is easy to verify that the number of solutions of the equation (\ref{eq:Sumation}) is given by:

\begin{equation}
\label{eq:mfactor}
m=\frac{(L+|I|-2L-1)!}{(|I|-2L)!(L-1)!}
\end{equation}

Note that $m$ increases very rapidly with $L$ and $|I|$. For example, if $|I|=100$ and $L=5$ we will generate $m=3.049.501$ solutions, and we will reach $m=40.430.556.376$ solutions if $|I|=1000$ and $L=5$. It is important to remark that, in the case of an exhaustive procedure, we will need to generate and also evaluate the variance (see equation \ref{eq:Varianc2}) for each one of these solutions.

Then, trying to reduce the potentially high number of arithmetic operations needed to obtain the optimal solution using the exhaustive enumerating process, we will instead solve the optimal stratification problem with proportional allocation by applying an algorithm based on the concept of minimum path in graphs.

Initially, we will translate the problem and its solution into a Graph Theory language \cite{Ahuja}, using the definitions of direct path and minimum path as follows.

Let $G=(V,A)$ a direct graph with  $|V|$  nodes,  $|A|$ arcs, and a cost $c_{ij} $ associated with each arc $(i,j)$  in $G$. 

Definition 1:  A directed path in $G$ is a directed walk without any repetition of nodes. In others words, a directed path has no backward arcs. The directed path is represented by a sequence of distinct nodes $i_{1} -i_{2} -...-i_{w} $, where   $\{ i_{1} ,i_{2} ,...,i_{w} \} \in V$. 

Definition 2:  Consider two nodes $i_{s} $ and $i_{d} $ of $G$. The minimal directed path between $i_{s} $ and  $i_{d} $  is a path whose sum of costs on the arcs in the path is minimal.

In order to make a correspondence of G with (part of) the solutions for the stratification problem, we associate to each index of the set $I$(see the beginning of this section) the nodes of the graph $G=(V,A)$, that means $V=I$. Moreover, to each arc$(i,j)\in A$ we associate a cost $c_{ij} $ given by $\frac{N}{n} .(1-\frac{n}{N} ).N_{h} .S_{hy}^{2} $, where $S_{hy}^{2} $ is the variance of the observations of  $Y$  that correspond to the observations of  $X$  which are associated to the indexes $\{ i$, $i+1$, $i+2,...,j-1\} \in I$.

Consequently, the values $N_{h} $ are also well defined for each arc (table 2). Each arc $(i,j)\in A$ corresponds to the indexes $\{ i$, $i+1$, $i+2,...,j-1\} \in I$ ; because of that an extra node $|I|+1$ has to be added to the set $V$.

In general, once defined the graph $G$ with its nodes and arcs, to resolve the problem of stratification will consist in finding a minimal path, from the node 1 to node $|I|+1$, with exactly $L$ arcs whose sum of costs $c_{ij} $ (variances) is a minimum. Thus, we must minimize equation (\ref{eq:Varianc2})

Taking into account these aspects and the fact that each stratum has to be defined by at least two observations (in order to avoid an indetermination when calculating $S_{hy}^{2} $ ), we can establish the following rules so as to determine each arc $(i,j)$ into $G$:

$\bullet$ The first stratum has to be defined by an arc $(i,j)$ which has a node $i=1$ and finishes at maximum in node $j=|I|+1-2(L-1).$ 

$\bullet$
The last stratum has to be defined by an arc $(i,j)$ which begins at least at the node $i=2.L-1$  and finishes exactly at the node $j=|I|+1-2(L-L)=|I|+1.$ 

$\bullet$
The intermediate strata $h$  ($1<h<L$) can be defined from the arcs$(i,j)$, such that  $2.h-1\le i\le |I|-1-2.(L-h)$   and   $i+1\le j\le |I|+1-2.(L-h)$.

$\bullet$
If  $j-i=1$ , the arc $(i,j)$ does not exist and then its cost is defined as $c_{ij} =+\infty $

Considering these rules, the total number of arcs in the first ($j-i\ne 1$), last and intermediary strata are given respectively by:

\begin{equation}
\label{eq:Total}
T_{1} =S, T_{L} =S, T_{h} =\frac{S^{2} +S}{2} , S=|I|-2L+1  
\end{equation}

Thus, the total  $T$  of arcs built for application of the minimum path algorithm is:

\begin{equation} 
\label{eq:General} 
T=T_{1} +T_{L} +(L-2).T_{h} =2.S+(L-2).(\frac{S^{2} +S}{2} ) 
\end{equation} 

In figure 1, we present a short example of a graph for a population where  $|X|=8$ and  $|I|=8$ and for  $L=3$. In table 2, we have all the possible arcs of the graph $G$ associated to the stratification problem, adopting the rules defined previously.

\begin{table}[!ht]
\begin{center}
{\scriptsize
\begin{tabular}{p{0.7in} p{0.5in} p{0.5in} p{3.5in}} \hline 
\textit{Stratum  $h$} & \textit{Node $i$  } & \textit{Node  $j$ } & \textit{Arcs $(i,j)$} \\ \hline
1 & 1 & 5 & \{(1,2), (1,3),(1,4),(1,5)\} \\ 
2 & 3 & 7 & \{(3,4),(3,5),(3,6),(3,7),(4,5),(4,6),(4,7),(5,6),(5,7)\} \\ 
3 & 5 & 9 & \{(5,9),(6,9),(7,9),(8,9)\} \\ \hline 
\end{tabular}
}
\end{center}
\caption{Arcs associated to the  graph of  figure 1 considering the rules of construction}
\end{table}

If we consider the sequence of arcs (path) defined by $\{(1,4), (4,7) e (7,9)\}$, the first strata would be defined by the indexes from 1 to 3, the second would have the indexes from 4  to  6 and the third, the indexes  7  and  8.

The example shown in figure 1 allows us to see the enumeration process of all paths from node 1 to $j=|I|+1=9$ containing $L=3$ arcs. So, a search to find a minimal path associated to a minimal variance can be carried on by considering just these paths.

\begin{figure}
\begin{center}
{\includegraphics[height=4 cm]{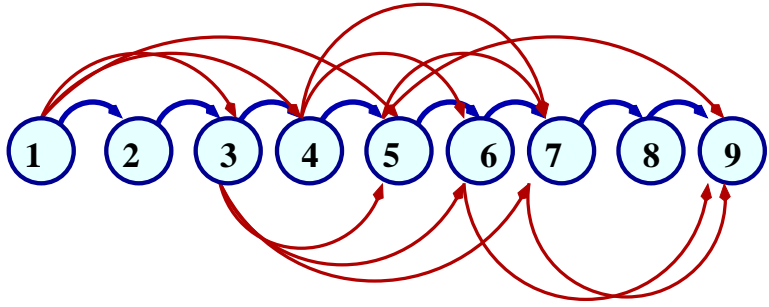} }
\end{center}
\caption{Example of a graph associated with the stratification problem}
\end{figure}

In order to find such a minimum path, we applied an extended Dijkstra's shortest path algorithm  (\cite{Ahuja} and \cite{Chen}). Dijkstra's algorithm finds shortest paths from the source node (vertex) \textit{s} to all other nodes in a network ($G$) with nonnegative arc costs. The extended version of the Dijkstra's algorithm takes into account the constraint that the number of arcs in the minimal path is a defined value $k$. Therefore, to solve the optimal stratification problem, we adjust the Dijkstra's algorithm in such a way to find a minimal path with exactly $L$ arcs. We call the resulting complete algorithm \textit{StratPath}.

In this algorithm, the input parameters are the number of strata, the size of the population, the sample size and the observations $X$ and $Y$, if a variable of study is considered.

Further details of the Dijkstra's algorithm and the Extended algorithm can be found in  \cite{Ahuja} and \cite{Chen}. What is important to emphasize is that, once the extended version of Dijkstra's algorithm gives optimal results (the shortest path) and the procedure we developed enumerates all the possible paths associated with all the possible solutions of the optimal stratification problem with proportional allocation, the \textit{StratPath} algorithm actually provides an optimal result.

Considering the application of the modified minimum path algorithm, the computational complexity associated to the resolution of this stratification problem stays in the order

\begin{equation}
\label{eq:complex}
O(2.S+(L-2).(\frac{S^{2} +S}{2} ))+O(L^{2} .(|I|+1)^{2} )
\end{equation}

In this expression, the first term is associated with number of operations (equation \ref{eq:General}) performed to generate all the arcs of the graph associated to the problem, and the second term corresponds to the number of operations performed by the minimum path algorithm  \cite{Chen} in order to obtain the solution of the problem.

\vspace{-0.8cm}

\section{Computational Results}

The algorithm Stratpath was implemented using C and Delphi language. The computational results reported in this section refer to a microcomputer IBM-PC with Pentium Processor IV (1.73 Ghz -- Dual Core) and 1GB of memory.

In order to assess the computational results provided by the algorithm proposed in this paper, consider three populations, extracted from \textit{PAM - Produ\c c\~ao Agr\'icola Municipal }(Municipal Agricultural Production), performed in 2005 by IBGE (Brazilian Central Statistical Office), which all have skewed positive distributions. These populations are associated to the total area of harvest in cities in the states of Rio Grande do Sul (population 1),  São Paulo (population 2)  and  Minas Gerais (population 3), considering the corn crop. These official data can be obtained from the internet in the following site:\\ (http://www.sidra.ibge.gov.br/bda/agric/).

We work under the assumption that the values of a study  variable $Y$ are equal those of the stratification variable $X$. Many other authors draw on this assumption, as Dalenius and  Hodges  \cite{Dalenius}, Hedlin \cite{Hedlin} and Lavall\'ee and Hidiroglou \cite{Lavallee}, among others.

In tables 3 and 4 it is possible to find results for $L$ = 4 and 5 strata. These tables present the values of  the populations size, set $|I|$, coefficients of variation $(CV=100.\sqrt{V(\widehat {X})} / \widehat {X})$, associated to estimates of totals of $X$ obtained by applying the \textit{StratPath} algorithm for each population, considering the minimization of the variance associated to proportional allocation (equation \ref{eq:Varianc2}). 

Moreover, they also present the population totals and  the sample sizes in each strata and the CPU times in seconds. We must emphasize that these sample sizes were obtained by using the equation \ref{eq:Proportional}.

\begin{table}[!ht]
\begin{center}
{\scriptsize
\begin{tabular}{p{0.8in} p{0.4in} p{0.4in} p{0.4in} p{0.4in} p{0.2in} p{0.3in} p{0.4in} p{0.3in} p{0.3in}} \hline 
\textit{Population  } & \textit{N} & $|I|$ & \textit{CPU} & \textit{CV} & \textit{Sizes} &  & Stratum &  &  \\
   & & & &  &  & 1 & 2  & 3 & 4  \\ \hline
\textit{}1 & 488 & 197 & 4 & 3.71 & \textit{Nh} & 262 & 168 & 50 & 8 \\ 
 &  & & &  & \textit{nh} & 54 & 34 & 10 & 2 \\ 
2 & 585 & 272 & 8 & 4.75 & \textit{Nh} & 447 & 104 & 24 & 10 \\ 
 &  &  & & & \textit{nh} & 76 & 18 & 4 & 2 \\ 
3 & 842 & 238 & 7 & 6.34 & \textit{Nh} & 680 & 130 & 16 & 16 \\ 
 &  &  & & & \textit{nh} & 81 & 15 & 2 & 2 \\ \hline 
\end{tabular}
}
\end{center}
\caption{Stratum Boundaries  with   L=4   and   n=100}
\end{table}

\begin{table}[!ht]
\begin{center}
{\scriptsize
\begin{tabular}{p{0.8in} p{0.4in}  p{0.4in} p{0.4in} p{0.4in} p{0.2in} p{0.3in} p{0.4in} p{0.3in} p{0.3in} p{0.3in}} \hline 
\textit{Population  } & \textit{N} & $|I|$ & \textit{CPU} & \textit{CV} & \textit{Sizes} &  & Stratum &  &  \\
   & & & &  &  & 1 & 2  & 3 & 4 & 5 \\ \hline
\textit{}1 & 488 & 197 & 5 & 3.32 &  \textit{Nh} & 245 & 150 & 74 & 14 & 5 \\ 
 &  & & &  & \textit{nh} & 50 & 31 & 15 & 3 & 1 \\ 

2 & 585 & 272 & 10 & 3.88 &  \textit{Nh} & 405 & 128 & 31 & 16 & 5 \\ 
 &  & & &  & \textit{nh} & 69 & 22 & 5 & 3 & 1 \\ 
3 & 842 & 238 & 8 & 4.94 &  \textit{Nh} & 633 & 135 & 43 & 16 & 15 \\
 &  & & &  & \textit{nh} & 75 & 16 & 5 & 2 & 2 \\ \hline
\end{tabular}
}
\end{center}
\caption{Stratum Boundaries  with   L=5   and   n=100}
\end{table}

In this paper, we presented an optimal algorithm for the problem of stratification by using the approach of minimal path in graphs.

This algorithm guarantees optimal strata boundaries, considering the assessment of the variance of a total estimator and a proportional allocation in strata and simple random sampling without replacement. This work can be extended to sampling with replacement ignoring the finite population correction.

As we can see from tables 3 and 4, CPU times were very low (up to 10 seconds), in such a way that the algorithm is hoped to run well even for reasonable greater instances.

In a future work, we intend to present a new algorithm that provides suitable boundaries for populations considering Neyman's allocation and other sample designs.


\begin{thebibliography}{10}
\expandafter\ifx\csname url\endcsname\relax
  \def\url#1{\texttt{#1}}\fi
\expandafter\ifx\csname urlprefix\endcsname\relax\def\urlprefix{URL }\fi

  

\bibitem{Ahuja}
K.R. Ahuja,  T.L. Magnanti, J.B. Orlin, \textit{ Networks Flows: Theory, Algorithms and Applications}, Prentice Hall, Inc. A Simon \& Schuster Company, 1993.

\bibitem{Chen}
S. Chen, K. Nahrstedt. On Finding Multi-Constrained Paths, \textit{ Proc. of IEEE International Conference on Communications}, Atlanta, GA, 1998, pp. 874-879.


\bibitem{Cochran}
W.G. Cochran, \textit{Sampling Techniques}, third ed., John Wiley \& Sons, New York, 1977.

\bibitem{Dalenius}
T. Dalenius, J.L. Hodges Jr., Minimum Variance Stratification, \textit{Skandinavisk Aktuarietidskrift}, 54 (1959) 88-101.

\bibitem{Gunning} 
P. Gunning, J.M. Horgan, A New Algorithm for the Construction of Stratum Boundaries in Skewed Populations, \textit{Survey Methodology}, 30 (2004) 159-166.

\bibitem{Hedlin}
D.A. Hedlin, A Procedure for Stratification by an Extended Ekman Rule, \textit{Journal
of Official Statistics}, 16 (2000) 15-29.

\bibitem{Lavallee}
P. Lavall\'ee, M.A. Hidiroglou, On the Stratification of Skewed
Populations, \textit{Survey Methodology (Statistics Canada)}, 14 (1988) 33-43.


\end{thebibliography}
\end{document}